\documentclass{article}
 
\usepackage{microtype}
\usepackage[utf8]{inputenc}
\usepackage{graphicx}
\usepackage{subfigure}
\usepackage{booktabs}
\usepackage{multirow}
\usepackage{hyperref}

\usepackage[accepted]{icml2025}
\usepackage{amsmath}
\usepackage{amssymb}
\usepackage{mathtools}
\usepackage{amsthm}
\usepackage[capitalize,noabbrev]{cleveref}
 
\theoremstyle{plain}

\theoremstyle{definition}

\theoremstyle{remark}

\usepackage[textsize=tiny]{todonotes}
 
\icmltitlerunning{Look Before You Steer: Geometry Predicts SAE Feature Steerability}
 
\begin{document}
 
\twocolumn[
\icmltitle{Look Before You Steer: Geometry Predicts SAE Feature Steerability}
 
\icmlsetsymbol{equal}{*}
 
\begin{icmlauthorlist}
\icmlauthor{Muhammad Khan}{equal,yyy}
\icmlauthor{Shlok Channawar}{equal,yyy}
\icmlauthor{Akshaj Gurugubelli}{yyy}
\icmlauthor{Girish Gupta}{yyy}
\icmlauthor{Aditya Shah}{yyy}
\end{icmlauthorlist}
 
\icmlaffiliation{yyy}{Algoverse AI Research}
\icmlcorrespondingauthor{Muhammad Khan}{Mo.aayuan.khan@gmail.com}
\icmlcorrespondingauthor{Shlok Channawar}{sfc5963@psu.edu}
\icmlcorrespondingauthor{Akshaj Gurugubelli}{akshajg9@gmail.com}
\icmlcorrespondingauthor{Girish Gupta}{girish@algoverseairesearch.org}
\icmlcorrespondingauthor{Aditya Shah}{aditya@algoverseairesearch.org}
\icmlkeywords{Machine Learning, ICML}
 
\vskip 0.3in
]
 
\printAffiliationsAndNotice{}
\def\thefootnote{*}\footnotetext{Equal contribution.}\def\thefootnote{\arabic{footnote}}
 
\begin{abstract}
Steering with SAE features requires per-feature coefficient tuning, which currently
demands intervention sweeps. We ask whether properties of the SAE itself, computable
before any forward pass, predict which features will be cheap or expensive to steer.
We show that variation in SAE feature steerability is partially predicted
by decoder-space geometry: neighbor density and maximum cosine similarity to
nearby decoder directions --- both computable from the SAE weight matrix
before any intervention --- rank features by how much steering they require for a
fixed behavioral effect ($\rho$ up to $-0.546$, $p < 10^{-6}$,
AUROC $0.610$--$0.822$ across conditions; the signal is rank-based, consistent with
grid discreteness). This geometry--steerability relationship replicates across two
Gemma-2 model scales (2B and 9B), two SAE widths (16K and 65K), and is detectable
cross-architecturally on Llama-3.1-8B-Instruct ($\rho = -0.266$, $n = 300$).
On Qwen3-8B with BatchTopK SAEs, geometry predicts whether a feature is
steerable at all but not the continuous ordering among responsive features,
revealing a boundary condition tied to SAE training regime.
The signal weakens at deep proportional layer depth in both models, where the
cost of steering exceeds our intervention budget --- a consistent depth
boundary. These results provide preliminary evidence that pre-steering
geometry can partially inform coefficient selection, offering a path toward
screening features for controllability before deployment.
\end{abstract}
 
\section{Introduction}
\label{sec:intro}
 
Large language models encode a vast range of behaviors in high-dimensional
activation spaces. Sparse autoencoders (SAEs) have emerged as a powerful tool
for decomposing those spaces into interpretable features \citep{cunningham2023sparse,
bricken2023monosemanticity}, and SAE features can be steered by amplifying their
decoder directions through direct activation-space interventions
\citep{templeton2024scaling}. Feature steering is increasingly applied to
safety-relevant behaviors --- refusal modulation, bias mitigation, truthfulness
control --- making reliable, predictable intervention a practical necessity.
 
Despite this promise, feature steering remains poorly understood at the individual
feature level. Practitioners select candidate features by inspecting top-activating
tokens, then manually sweep steering coefficients and observe outcomes. This workflow
is expensive, inconsistent, and fundamentally retrospective; there is no principled
way to estimate, before intervening, how much force a feature will require.
This paper focuses on predicting intervention magnitude. (Measuring collateral risk
is deferred to future work.) Activation evidence alone is insufficient to determine
causal behavioral influence \citep{arad2025saesteering}, and no existing framework
provides a pre-intervention, feature-level characterization of either controllability
or off-target risk \citep{anthropic2024steering}.
 
We show that the variation in steering cost across features is partially predicted by decoder-space geometry.
Features embedded in dense decoder neighborhoods respond at smaller coefficients, while geometrically isolated features require larger coefficients to reach the same behavioral threshold. Formalizing \emph{steerability} as the minimum
steering coefficient $\alpha^*$ required to produce a fixed behavioral change, we
find that neighbor density and maximum cosine similarity to nearby decoder
directions --- both computable from the SAE weight matrix before any intervention
is applied --- consistently predict $\alpha^*(f)$ with Spearman $\rho$ up to
$-0.546$ ($p < 10^{-6}$).
 
This signal replicates across Gemma-2 model scales and SAE widths, is detectable
on Llama-3.1-8B-Instruct ($\rho = -0.266$), and is stronger in 9B than 2B
at matched layer index. On Qwen3-8B with BatchTopK SAEs, geometry predicts binary
steerability but not the continuous ordering among responsive features. The signal
weakens at deep proportional layer depth --- a consistent boundary condition.
Co-activation correlation carries no independent predictive signal once activation
sparsity is accounted for.
 
These results suggest that pre-steering geometry can partially inform coefficient
selection. A practitioner with access to the SAE decoder weight matrix can identify geometrically dense features before running a single steering experiment, prioritizing them as more responsive intervention targets. This is a step toward
transforming feature steering from ad-hoc exploration into a principled, partially
predictable procedure --- with implications for screening intervention candidates
in safety-critical settings.
 
Our contributions are threefold. First, we formalize SAE feature steerability as the minimum coefficient required to reach a fixed behavioral threshold. Second, we show that simple decoder-geometry metrics rank-order steering cost across Gemma-2, Llama-3.1, and Qwen3 conditions, with clear boundary cases. Third, we show that co-activation, as measured on a small task corpus, contributes little reliable predictive signal. We focus on whether pre-steering geometry predicts the coefficient required to produce a fixed behavioral effect; off-target analysis and safe coefficient rules are deferred to future work.
 
\section{Related Work}
\label{sec:related}
 
Sparse autoencoders decompose activations into sparse feature dictionaries and have been used to identify interpretable directions in language models \citep{cunningham2023sparse,bricken2023monosemanticity,lieberum2024gemma}. Activation-space steering methods show that modifying internal representations can change model behavior \citep{li2023iti,zou2023repeng,panickssery2024caa,turner2024actadd}, and recent SAE-steering work emphasizes that feature selection matters \citep{arad2025saesteering,anthropic2024steering}. However, existing work largely selects and tunes features retrospectively. In parallel, work on representation geometry and superposition suggests that local structure in activation space may affect intervention behavior \citep{park2024geometry,li2025tracing,elhage2022superposition}, but this connection has not been tested as a feature-level predictor of steering cost.
 
\section{Problem Formulation}
\label{sec:formulation}
 
Let $f$ be an SAE feature with decoder direction $\mathbf{v}_f$. We steer by
adding a scaled copy of $\mathbf{v}_f$ to the model's residual-stream activation
at a chosen layer:
\begin{equation}
    \mathbf{h}' = \mathbf{h} + \alpha \cdot \mathbf{v}_f,
    \label{eq:steer}
\end{equation}
where $\alpha$ is the steering coefficient.
 
Intuitively, geometry predicts steerability because steering is never perfectly
targeted. If decoder directions $\mathbf{v}_f$ and $\mathbf{v}_g$ have cosine
similarity $c$, then adding $\alpha \cdot \mathbf{v}_f$ to the residual stream
also shifts $\mathbf{v}_g$'s reconstruction by approximately $\alpha \cdot c$.
Features in dense decoder neighborhoods therefore recruit nearby correlated 
directions when steered, and when those neighbors share the target behavior 
this co-recruitment lowers the coefficient required to reach threshold; 
isolated features, with no aligned neighbors to amplify the push, require 
larger $\alpha$ to reach the same behavioral threshold.
 
\begin{figure}[t]
  \centering
  \includegraphics[width=\columnwidth]{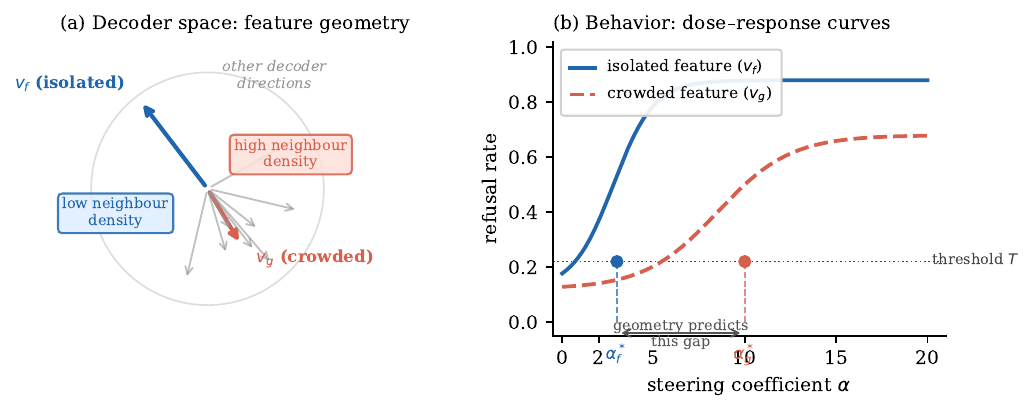}
  \caption{Illustration of the geometry--steerability relationship.
  (a) Two SAE features in decoder space. The isolated feature ($v_f$, blue) has
  few nearby decoder directions; the crowded feature ($v_g$, red) sits in a dense
  cluster. Grey arrows are other features in the same dictionary. (b) Steering each
  feature: the crowded one crosses the refusal threshold $T$ at a much smaller
$\alpha^*$ than the isolated one.
  weight matrix before any steering --- predicts this gap. Curves are schematic;
  empirical results are shown in Figure~\ref{fig:scatter_9b_l20}.}
  \label{fig:intuition}
\end{figure}
 
Let $B(\alpha, f)$ denote a scalar on-target behavior score under coefficient
$\alpha$. We define steerability as the minimum coefficient that produces a fixed
behavioral improvement:
\begin{equation}
    \alpha^*(f) = \min\bigl\{\alpha : B(\alpha, f) - B(0) \ge T\bigr\},
    \label{eq:astar}
\end{equation}
where $T$ is a pre-registered behavioral threshold. If no tested coefficient
reaches the threshold, $\alpha^*(f)$ is right-censored:
$\alpha^*(f) > \alpha_{\max}$.
 
\subsection{Pre-Steering Feature Metrics}
\label{sec:metrics}
 
For each feature $f$, we compute three metrics from the SAE decoder weight matrix
and a reference corpus before any steering is applied.
 
The first metric is \emph{max cosine similarity}:
$\max_{g \ne f} \cos(\mathbf{v}_f, \mathbf{v}_g)$. High values indicate the
feature lies near at least one other decoder direction.
 
The second is \emph{neighbor density}, defined as the mean cosine similarity to
the $k$ nearest decoder neighbors (default $k{=}50$). For the 65K condition, where
the denser decoder space motivates examination of distributional shape beyond the
mean, we additionally report \emph{density-$\tau$} --- the Kendall $\tau$ rank
correlation between a feature's cosine similarities to its $k$ nearest neighbors
and their rank order --- as a non-parametric robustness check on the mean-based
estimate.
 
The third is \emph{co-activation correlation}, the mean Pearson correlation between
$f$'s activations and those of its top-$k$ co-active features on a reference
corpus. This captures functional entanglement beyond pure geometry.
 
\subsection{Well-Identified Conditions}
\label{sec:wellidentified}
 
We define a condition as \emph{well-identified} if it satisfies two criteria
evaluated prior to analysis: (1) censoring rate below 50\% (i.e., at least half
of features reach threshold within the tested coefficient range), and (2) no
coherence collapse in the steerability distribution (i.e., the dose-response curve
is not disrupted by degenerate completions spanning the majority of steering grid
points). Conditions failing either criterion are reported for completeness but
excluded from primary inference. Under this definition,two of the six Gemma-2 conditions are
not well-identified: 2B layer 24 (coherence collapse) and 9B layer 36 (46\%
censoring, borderline; geometry predicts among uncensored features but primary
inference is attenuated).
 
\section{Methods}
\label{sec:methods}
 
\subsection{Phase 1: Feature Selection and Characterization}
 
We select features through a two-stage procedure applied identically across all
experimental conditions. First, contrast scoring: for each SAE feature, we compute
the difference in activation frequency between task-relevant prompts
(SALADBench-derived) and neutral control text, ranking by a composite contrast
score and retaining the top 300. Feature dictionary sizes vary by condition: 16,384
(GemmaScope 16K), 65,536 (GemmaScope 65K), 131,072 (Llama-3.1-8B BatchTopK), and
32,768 (Qwen3-8B BatchTopK). Second, output-score filtering: we apply an
Arad-style causal relevance check \citep{arad2025saesteering} that measures whether
each feature's decoder direction shifts the model's output distribution when
clamped. Features failing a minimum effect threshold ($\delta \geq 0.01$) are
discarded, yielding final sets of 75--300 features per condition (75 for 9B layer
20; 100 for all other Gemma-2 conditions; 113 for Qwen3-8B; 300 for Llama-3.1-8B,
where all contrast-selected features passed the causal filter).
 
Pre-steering metrics (Section~\ref{sec:metrics}) are computed once for this
filtered set.
 
\subsection{Phase 2: Measuring Steerability}
 
For each selected feature and each coefficient in a grid
$\alpha \in \{0, 0.25, 0.5, 1, 2, 3, 5, 10, 20\}$, we generate completions on a
fixed prompt set of 100 SALADBench-derived prompts and compute the on-target
behavioral score $B(\alpha, f)$. Note that we apply only positive steering
($\alpha > 0$), targeting amplification of refusal-relevant features; the
adversarially relevant direction (suppression via $\alpha < 0$) is a planned
extension. We extract $\alpha^*(f)$ per Eq.~\eqref{eq:astar} with threshold
$T = 0.10$ (absolute shift in refusal rate). This threshold was chosen as a
practically meaningful shift: a 10-point increase in refusal rate exceeds the
variability observed across prompt phrasings in pilot runs ($\pm 3$--$4$ points)
and corresponds to a detectable behavioral change while remaining below the regime
of complete output saturation. Sensitivity to
$T \in \{0.05, 0.10, 0.15, 0.20\}$ is reported in Appendix~\ref{app:threshold}.
A reference coefficient $\alpha_{\text{ref}} = -20$ is used for baseline estimation
only; all behavioural measurements use positive $\alpha$. Features not reaching
threshold within the tested range are right-censored at
$\alpha^*(f) > \alpha_{\max}$.
 
To guard against degenerate completions, we apply a coherence filter that removes
rows where the model output collapses under steering. A completion is flagged as
degenerate if it satisfies any of the following: (a) the output consists of fewer
than five tokens (token collapse), (b) any token type repeats more than 15
consecutive times (repetition collapse), or (c) the perplexity under the unsteered
model exceeds $e^{5}$, equivalently a mean token log-probability below $-5$
(incoherent fluency). Flagged rows are excluded from behavioral score computation.
At layer 20, approximately 24\% of rows were filtered as degenerate.
 
\subsection{Phase 3: Predictability Analysis}
 
We test whether pre-steering metrics predict $\log \alpha^*(f)$ using:
(1) Spearman rank correlations between each metric and $\log \alpha^*(f)$;
(2) cross-validated $R^2$ from ridge regression using all three metrics;
(3) AUROC for binary classification of steerable ($\alpha^* \leq 3$) versus
hard-to-steer features; (4) permutation-based null tests ($n = 1{,}000$ shuffles)
with bootstrap confidence intervals; and (5) leave-one-metric-out ablations to
assess the marginal contribution of each predictor. Censored features are included
in Spearman correlations using standard rank-based handling of ties at the boundary
and excluded from $R^2$ regression to avoid artificial ceiling effects.
 
\section{Experimental Setup}
\label{sec:setup}
 
\subsection{Model and SAE}
 
Table~\ref{tab:model_sae} summarises all experimental conditions. Our primary
experiments use Gemma-2-2B-it and Gemma-2-9B-it with GemmaScope residual-stream
SAEs \citep{lieberum2024gemma}, covering layers 20, 22, and 24 (2B) and layers 20
and 36 (9B), with both 16K and 65K dictionary widths at 2B layer 20. For
cross-architecture generalization we run the full pipeline on Llama-3.1-8B-Instruct
with a BatchTopK SAE \citep{bussmann2024batchtopk,karvonen2024llama} at layer 23
(72\% depth, 131K features), and on Qwen3-8B with a BatchTopK SAE
\citep{karvonen2024qwen} at layer 18 (Appendix~\ref{app:qwen}). Steering is applied
at the residual stream post-attention; all experiments use positive steering only
($\alpha > 0$).
 
\begin{table*}[!t]
\caption{Experimental conditions.}
\label{tab:model_sae}
\vskip 0.1in
\begin{center}
\begin{small}
\begin{tabular}{llccc}
\toprule
Model & SAE & Layer(s) & Dict. & Rel. depth \\
\midrule
Gemma-2-2B & GemmaScope 16K & 20, 22, 24 & 16K & 77\%, 85\%, 92\% \\
Gemma-2-2B & GemmaScope 65K & 20 & 65K & 77\% \\
Gemma-2-9B & GemmaScope 16K & 20, 36 & 16K & 48\%, 86\% \\
Llama-3.1-8B & BatchTopK & 15 & 131K & 47\% \\
Llama-3.1-8B & BatchTopK & 23 & 131K & 72\% \\
Qwen3-8B & BatchTopK & 18 & 32K & 50\% \\
\bottomrule
\end{tabular}
\end{small}
\end{center}
\vskip -0.1in
\end{table*}
 
\subsection{Datasets}
 
Geometric pre-steering metrics (neighbor density and maximum cosine similarity) are
computed directly from the SAE decoder weight matrix $W_{\text{dec}}$ and require
no corpus. Co-activation correlation is estimated from baseline forward passes on
99 SALADBench-derived prompts, the same pool used for feature selection.
 
We measure refusal behavior using 100 prompts drawn from SALADBench
\citep{li2024saladbench}, covering safety-relevant query categories. The on-target
score $B(\alpha, f)$ is the mean refusal rate across prompts under steering,
measured by a keyword-based refusal classifier. The prompt set is generated once
via a deterministic preparation script (seed 42) and frozen across all runs to
ensure comparability across all experimental conditions.
 
Classifier reliability is reported in Appendix~\ref{app:scorer}. Off-target
evaluation is reserved for future work.
 
\subsection{Feature Selection Pipeline}
 
The two-stage selection procedure (Section~\ref{sec:methods}) is applied
identically across all conditions; see Section~\ref{sec:methods} for full details
and feature counts.
 
\subsection{Baselines and Controls}
 
We include two controls. A permutation null shuffles metric-to-feature assignments
($n = 1{,}000$) and repeats the correlation analysis to establish chance-level
$\rho$ and $R^2$. A metric ablation removes each metric individually and assesses
the change in cross-validated $R^2$ and AUROC.
 
\section{Results}
\label{sec:results}
 
\subsection{Feature Selection Summary}
 
Starting from condition-specific SAE dictionaries (16,384 to 131,072 features
depending on the SAE), contrast-based selection retained the top 300 by composite
activation contrast score. Output-score filtering then discarded features failing
the causal relevance threshold ($\delta \geq 0.01$), yielding final sets of
75--300 features per condition. For Llama-3.1-8B-Instruct, all 300 passed the
causal filter, consistent with a richer pool of refusal-relevant directions in the
larger dictionary. Per-condition geometry statistics are reported in
Appendix~\ref{app:geom_stats}.
 
Co-activation correlation statistics are interpreted with caution: the metric is
heavily zero-inflated due to features that never activate on the 99-prompt reference
corpus (Section~\ref{sec:h1}), making summary statistics unreliable as a
characterization of functional entanglement.
 
\subsection{Steerability Distribution}
 
\begin{table*}[!t]
\caption{Steerability distribution across experimental conditions.
$N_{\text{cens}}$ is the number of right-censored features
($\alpha^* > \alpha_{\max}$). Mean and median $\alpha^*$ are computed over
uncensored features only. Relative depth is layer index divided by total layers.
Runs is the number of independent pipeline runs completed for each condition.
$^\dagger$Distribution collapsed due to coherence collapse at high $\alpha_{\max}$;
excluded from primary inference per Section~\ref{sec:wellidentified}.}
\label{tab:steerability}
\vskip 0.15in
\begin{center}
\begin{small}
\begin{sc}
\begin{tabular}{llcccccc}
\toprule
Model & Condition & Rel.\ depth & Runs & $N$ & $N_{\text{cens}}$ (\%) &
Mean $\alpha^*$ & Median $\alpha^*$ \\
\midrule
\multirow{4}{*}{2B}
 & L20 16K & 77\% & 3 & 100 & 12 (12\%) & 3.39 & 3.0 \\
 & L22 16K & 85\% & 3 & 100 & 27 (27\%) & 3.84 & 3.0 \\
 & L24 16K$^\dagger$ & 92\% & 3 & 100 & --- & --- & --- \\
 & L20 65K & 77\% & 4 & 100 & 11 (11\%) & 3.39 & 3.0 \\
\midrule
\multirow{2}{*}{9B}
 & L20 16K & 48\% & 3 & 75  &  2 (3\%)  & 4.25 & 3.0 \\
 & L36 16K & 86\% & 2 & 100 & 46 (46\%) & 4.83 & 5.0 \\
\midrule
\textsc{Llama-3.1-8B} & L15 \textsc{BatchTopK} & 47\% & 1 & 300 & 0 (0\%) &
6.11 & 5.0 \\
\textsc{Llama-3.1-8B} & L23 \textsc{BatchTopK} & 72\% & 1 & 300 & 0 (0\%) &
6.71 & 5.0 \\
\bottomrule
\end{tabular}
\end{sc}
\end{small}
\end{center}
\vskip -0.1in
\end{table*}
 
Table~\ref{tab:steerability} summarizes the steerability distribution across all
experimental conditions. For the 16K SAE at layer 20 (2B), 88 of 100 features
were steerable under positive steering (12\% censored), with $\alpha^*(f)$ ranging
from 2 to 10 (mean 3.39, median 3.0). The distribution was concentrated at
$\alpha^* \in \{2, 3, 5\}$, with 23, 39, and 25 features respectively. At layer
22 (2B), censoring increased to 27\% (mean 3.84, median 3.0), consistent with the
cost of steering growing at deeper layers under the same coefficient grid. For the
65K SAE at layer 20 (2B), censoring was similar to the 16K condition at 11\%
(mean 3.39, median 3.0).
 
The 9B model at layer 20 was dramatically more steerable: only 2 of 75 features
were censored (2.7\%), the lowest censoring rate across all Gemma-2 conditions,
with mean $\alpha^* = 4.25$ and median 3.0. This suggests that at the same
absolute layer depth, refusal-relevant features in the larger model respond more
readily to steering interventions. At layer 36 (9B), censoring rose sharply to
46\%: only 54 of 100 features reached threshold within the coefficient grid
(mean 4.83, median 5.0), with the majority exceeding our intervention budget.
Inspection of raw rollouts confirms that censored features did exhibit behavioral
change under steering: refusal rates dropped substantially at $\alpha = -20$,
indicating that these features are not unresponsive but rather that the cost of
reaching threshold exceeds the range of our current grid.
 
For Llama-3.1-8B-Instruct at layer 23 (BatchTopK, 131K), zero features were
censored ($\alpha^*(f) \leq \alpha_{\max}$ for all 300), indicating that all
features eventually reach threshold within the coefficient grid, though at higher
mean $\alpha^*$ than either Gemma-2 variant. The majority of features cluster at
$\alpha^* = 5$, with mean 6.71 and median 5.0.
 
Two depth-dependent patterns emerge, one in each model, which we discuss in
Section~\ref{sec:depth_boundary}.
 
\subsection{Pre-Steering Metrics Predict Steerability (RQ1)}
\label{sec:h1}

Our keyword classifier achieves 57.5\% recall (Appendix~\ref{app:scorer}), systematically undercounting soft refusals; reported $\alpha^*$ values are therefore likely overestimates and Spearman correlations likely attenuated, meaning the true geometry--steerability relationship is probably stronger than reported.

Table~\ref{tab:correlations} reports Spearman correlations between each
pre-steering metric and $\log \alpha^*(f)$ across all conditions.
Table~\ref{tab:classification} reports cross-validated $R^2$, AUROC, and
permutation-null results from Phase 3. We emphasize that the predictive signal is
rank-based; ridge regression on $\log \alpha^*$ explains little additional
variance (CV $R^2$ near zero in Table~\ref{tab:classification}), consistent with
the discreteness of our coefficient grid. Geometry rank-orders features by
steerability difficulty but does not predict the magnitude of $\alpha^*$.
 
\begin{table*}[t]
\caption{Spearman correlations between pre-steering metrics and
$\log \alpha^*(f)$. $r_{\text{full}}$ includes censored features (rank-based
handling of ties at boundary); $r_{\text{nocens}}$ excludes them. $n$ is the
number of uncensored features used for $r_{\text{nocens}}$. For Llama-3.1-8B,
$r_{\text{full}} = r_{\text{nocens}}$ since censoring is zero. Qwen3-8B results
are reported in Appendix~\ref{app:qwen}.}
\label{tab:correlations}
\vskip 0.15in
\begin{center}
\begin{small}
\begin{tabular}{llccccr}
\toprule
Condition & Metric & $r_{\text{full}}$ & $p_{\text{full}}$ &
$r_{\text{nocens}}$ & $p_{\text{nocens}}$ & $n$ \\
\midrule
\multirow{3}{*}{2B L20 16K (3 runs)}
 & max\_cosine       & $-0.192$ & $0.055$             & $-0.299$ & $0.005$            & 88 \\
 & neighbor\_density & $-0.206$ & $0.039$             & $-0.288$ & $0.006$            & 88 \\
 & coactivation      & $+0.030$ & $0.765$             & $+0.184$ & $0.087$            & 88 \\
\midrule
\multirow{3}{*}{2B L22 16K (3 runs)}
 & max\_cosine       & $-0.293$ & $0.003$             & $-0.425$ & $0.0002$           & 73 \\
 & neighbor\_density & $-0.222$ & $0.026$             & $-0.512$ & $3.6\times10^{-6}$ & 73 \\
 & coactivation      & $+0.147$ & $0.145$             & $+0.147$ & $0.214$            & 73 \\
\midrule
\multirow{3}{*}{2B L24 16K (3 runs)}
 & max\_cosine       & $-0.135$ & $0.181$             & $-0.150$ & $0.172$            & 54 \\
 & neighbor\_density & $-0.149$ & $0.138$             & $-0.125$ & $0.256$            & 54 \\
 & coactivation      & $+0.100$ & $0.321$             & $+0.132$ & $0.233$            & 54 \\
\midrule
\multirow{4}{*}{2B L20 65K (4 runs)}
 & max\_cosine       & $-0.329$ & $0.0008$            & $-0.404$ & $0.0001$           & 89 \\
 & neighbor\_density & $-0.394$ & $0.0001$            & $-0.435$ & ${<}0.0001$        & 89 \\
 & density\_$\tau$   & $-0.410$ & ${<}0.0001$         & ---      & ---                & 100 \\
 & coactivation      & $+0.079$ & $0.440$             & $+0.151$ & $0.160$            & 89 \\
\midrule
\multirow{3}{*}{9B L20 16K (3 runs)}
 & max\_cosine       & $-0.461$ & $3.2\times10^{-5}$  & $-0.490$ & $1.1\times10^{-5}$ & 73 \\
 & neighbor\_density & $-0.523$ & $1.5\times10^{-6}$  & $-0.546$ & $5.8\times10^{-7}$ & 73 \\
 & coactivation$^*$  & $+0.328$ & $0.004$             & $+0.313$ & $0.007$            & 73 \\
\midrule
\multirow{3}{*}{9B L36 16K (2 runs)}
 & max\_cosine       & $-0.284$ & $0.004$             & $-0.409$ & $0.002$            & 54 \\
 & neighbor\_density & $-0.258$ & $0.009$             & $-0.327$ & $0.016$            & 54 \\
 & coactivation      & $+0.027$ & $0.786$             & $+0.141$ & $0.311$            & 54 \\
\midrule
\multirow{3}{*}{Llama-3.1-8B L15 BatchTopK (1 run)}
 & max\_cosine       & $-0.266$ & $3.0\times10^{-6}$  & $-0.266$ & $3.0\times10^{-6}$ & 300 \\
 & neighbor\_density & $-0.188$ & $0.0011$            & $-0.188$ & $0.0011$           & 300 \\
 & coactivation      & $-0.124$ & $0.031$             & $-0.124$ & $0.031$            & 300 \\
\midrule
\multirow{3}{*}{Llama-3.1-8B L23 BatchTopK (1 run)}
 & max\_cosine       & $-0.163$ & $0.0046$            & $-0.163$ & $0.0046$           & 300 \\
 & neighbor\_density & $-0.220$ & $0.0001$            & $-0.220$ & $0.0001$           & 300 \\
 & coactivation      & $-0.134$ & $0.0202$            & $-0.134$ & $0.0202$           & 300 \\
\bottomrule
\end{tabular}
\end{small}
\end{center}
{\footnotesize $^*$Zero-inflated; see text.}
\vskip -0.1in
\end{table*}
 
\begin{table*}[t]
\caption{Predictability analysis: cross-validated $R^2$ (ridge regression on all
three metrics), AUROC (steerable $\alpha^* \leq 3$ vs.\ hard-to-steer), and
permutation-null 95th percentile $\rho$ across well-identified conditions.
Censored features excluded from $R^2$ and AUROC; included (rank-tied) in Spearman.
Llama-3.1-8B L23 is omitted; cross-architecture Spearman correlations for that
condition are reported in Table~\ref{tab:correlations}.}
\label{tab:classification}
\vskip 0.15in
\begin{center}
\begin{small}
\begin{tabular}{lcccccc}
\toprule
Condition & $n$ & CV $R^2$ & AUROC & Perm.\ null $\rho_{95}$ &
LOO: drop density & LOO: drop max cos \\
\midrule
2B L20 16K        & 88 &  0.068   & 0.610 & 0.168 &  0.054   &  0.075   \\
2B L22 16K        & 73 &  0.047   & 0.736 & 0.194 & $-$0.045 &  0.109   \\
2B L20 65K        & 89 & $-$0.005 & 0.686 & 0.172 & $-$0.048 &  0.022   \\
9B L20 16K        & 73 &  0.025   & 0.822 & 0.197 & $-$0.028 &  0.053   \\
9B L36 16K$^\dag$ & 54 & $-$0.698 & 0.723 & 0.225 & $-$0.607 & $-$0.742 \\
Llama L15 131K     & 300 & --- & 0.653 & --- & --- & --- \\
\bottomrule
\end{tabular}
\end{small}
\end{center}
{\small $^\dag$ High censoring (46\%) inflates ridge regression instability;
AUROC and Spearman remain interpretable among the 54 uncensored features.}
\end{table*}
 
Neighbor density and maximum cosine similarity consistently predict
$\log \alpha^*(f)$ across all well-identified conditions. Three patterns emerge
across the full experimental matrix.
 
First, the signal is stronger in 9B than 2B at matched layer index within the
Gemma-2 family. At layer 20, neighbor density $\rho$ increases from $-0.288$ (2B,
$n=88$) to $-0.546$ (9B, $n=73$, $p = 5.75 \times 10^{-7}$) among uncensored
features, suggesting that larger models develop more geometrically structured SAE
features whose isolation more reliably predicts steerability. The 9B result is the
strongest in the entire experimental dataset.
 
Second, the signal collapses at the deepest tested layer in each model. In the 2B,
neighbor density is strongest at layer 22 ($\rho = -0.512$, 85\% depth) and
collapses to null at layer 24 (92\% depth). In the 9B, it drops from
$\rho = -0.546$ at layer 20 (48\% depth) to $\rho = -0.327$ at layer 36 (86\%
depth), where 46\% censoring indicates most features lie beyond the range of
intervention. Geometry still predicts among the steerable subset at 9B L36 (both
metrics $p < 0.02$), but the practical reach of the framework narrows substantially
at deep layers.
 
Third, the signal is detectable cross-architecturally on Llama-3.1-8B-Instruct.
 
\paragraph{Cross-architecture generalization.}
On Llama-3.1-8B-Instruct we ran the full pipeline at two layers. At layer 15
(47\% depth, $n=300$, 0\% censored), neighbor density $\rho = -0.188$
($p = 0.001$) and max cosine $\rho = -0.266$ ($p < 0.0001$) --- stronger than
the layer 23 result (neighbor density $\rho = -0.220$, max cosine $\rho = -0.163$),
consistent with the depth-dependent pattern observed within Gemma-2: shallower
layers show stronger geometric signal. At layer 23 (72\% depth, $n=300$, 0\%
censored), the signal attenuates but remains statistically significant (neighbor
density $\rho = -0.220$, $p = 1.24 \times 10^{-4}$; max cosine $\rho = -0.163$,
$p = 4.63 \times 10^{-3}$). The consistent negative direction and statistical
significance across both layers and 300 features constitute a positive
cross-architecture replication of the core geometric hypothesis. Effect sizes are
smaller than in Gemma-2 ($\rho = -0.266$ vs.\ $-0.546$ at 9B L20), which we
attribute primarily to the BatchTopK training regime enforcing batch-level sparsity
rather than per-token thresholds, potentially decoupling decoder geometry from
functional behavior to a greater degree. Both Llama results are based on single
pipeline runs and should be treated as preliminary; multi-run replication is
planned.

To probe the mechanism underlying the geometry--steerability relationship, 
we computed the mean cosine similarity of each feature's $k=50$ decoder 
neighbors to the mean refusal direction (the centroid of selected feature 
decoder vectors). Neighbor refusal alignment correlates negatively with 
$\alpha^*(f)$ ($\rho = -0.450$, $p = 0.0001$, $n = 73$): easy-to-steer 
features have neighbors approximately $3.4\times$ more aligned with the 
refusal direction than hard-to-steer features (mean alignment $0.129$ vs.\ 
$0.038$), consistent with co-recruitment of refusal-adjacent neighbors 
amplifying the behavioral effect.
 
The 65K SAE replication at layer 20 (2B) confirms that the signal reflects
\emph{relative} geometric crowding rather than absolute geometry: despite the 65K
SAE having substantially higher absolute cosine values (max cosine mean 0.547
vs.\ 0.392 for 16K), effect sizes are comparable ($\rho = -0.435$ vs.\ $-0.288$
for neighbor density), and the signal survives a fourfold increase in dictionary
size. A robustness check at layer 22 dropping the two outlier features at
$\alpha^* = 10$ yielded $\rho = -0.470$ ($p = 3.6 \times 10^{-5}$), confirming
results are not driven by boundary cases.
 
\paragraph{Co-activation correlation.}
Co-activation correlation requires careful interpretation. In five of six conditions
it is null ($p > 0.08$). At 9B L20, the full-sample correlation appears significant
($r_{\text{full}} = +0.328$, $p = 0.004$), but this is attributable to
zero-inflation: 59 of 75 features have activation frequency zero on the 99-prompt
reference corpus, and all zero-frequency features have coactivation exactly zero by
construction. Within the 16 features that actually activate on the reference corpus,
the coactivation--$\alpha^*$ correlation is $p = 0.38$ (null). At Llama-3.1-8B
L23, co-activation correlation reaches marginal significance ($\rho = -0.134$,
$p = 0.020$), but the sign is negative (consistent with geometry), the effect is
smaller than both geometry metrics, and we treat this as suggestive rather than
conclusive given the single-run design. We conclude that co-activation correlation
as operationalized here --- mean Pearson correlation over a small task-specific
corpus --- is not a reliable pre-steering predictor of steerability, and that the
metric requires a larger or more targeted activation corpus to be interpretable.
Geometry appears to be the dominant predictive signal in our data; co-activation
contributes little independent information.
 
\begin{figure*}[t]
  \centering
  \includegraphics[width=\textwidth]{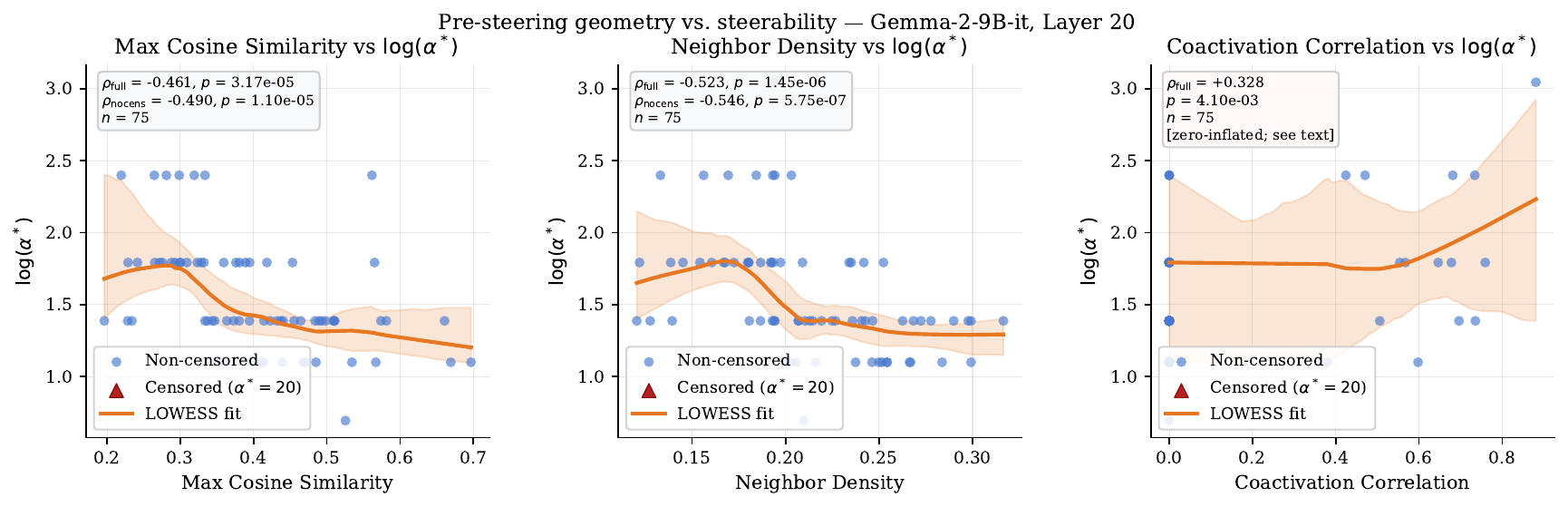}
  \caption{Pre-steering metrics versus log steerability for the 9B model at layer
  20 (strongest signal condition). Each point is one SAE feature; red triangles
  indicate right-censored features ($\alpha^* = 20$). Trend line shows LOWESS fit;
  shaded band is bootstrap 95\% CI. Neighbor density: $\rho = -0.546$,
  $p = 5.75 \times 10^{-7}$. Max cosine similarity: $\rho = -0.490$,
  $p = 1.10 \times 10^{-5}$. Co-activation correlation: zero-inflated; see text.}
  \label{fig:scatter_9b_l20}
\end{figure*}
 
\paragraph{Activation-frequency baseline.}
As an additional baseline, we tested whether behavioral proxies available prior to
steering predict $\alpha^*(f)$ comparably to geometric metrics. The change in
refusal rate at $\alpha = 1$ --- that is, $B(1, f) - B(0, f)$, the behavioral
shift at the smallest non-zero grid point, a forward-pass-only proxy requiring one
steered run per feature rather than a full sweep --- yields $\rho = -0.339$
($p = 0.001$, $n = 89$), weaker than both neighbor density ($\rho = -0.435$) and
max cosine similarity ($\rho = -0.404$). Larger coefficient responses ($\alpha = 5$,
$\alpha_{\text{ref}} = -20$) are not significant predictors. Base refusal rate is
constant across features and carries no signal. This suggests that decoder-space
geometry provides predictive information beyond what is recoverable from behavioral
measurements at low steering coefficients.
 
\paragraph{Off-target effects (RQ2 and RQ3).}
Whether harder-to-steer features also produce greater off-target effects (RQ2), and
whether geometrically dense features pose structural risk even at low shared
coefficients (RQ3), are planned extensions. The geometry signal identified here
provides a natural basis for both: if crowded decoder neighborhoods predict intervention ease, 
they may also predict collateral risk. Empirical validation
is deferred to future work.
 
\subsection{Depth Boundary}
\label{sec:depth_boundary}
 
Two depth-dependent patterns emerge, one in each model. Here $\alpha_{\max}$ denotes
the residual-stream activation norm at the target layer, which scales the effective
force applied by a fixed steering coefficient. At layer 24 (2B, 92\% relative
depth), the steerability distribution collapsed entirely, which we attribute to
substantially higher $\alpha_{\max}$ values (${\sim}188$ vs.\ ${\sim}75$ at layer
22) causing the fixed coefficient grid to apply approximately $2.5\times$ more force
and inducing coherence collapse in the dose-response curve. At layer 36 (9B, 86\%
relative depth), the effect is partial: geometry still predicts steerability among
the 54 features that reach threshold, but the majority require steering coefficients
beyond our current grid. Both patterns are consistent with the cost of steering
growing as deep proportional layer depth approaches the output layer. We note that
the 46\% censoring at 9B L36 also attenuates our Spearman estimates by collapsing
rank variance among features tied at $\alpha^* = \alpha_{\max}$; the true
geometry--steerability relationship at this depth may be stronger than our reported
correlations indicate.
 
Within the Gemma-2 family, the pattern is consistent: Spearman $\rho$ for neighbor
density moves from $-0.288$ (2B L20, 77\% depth) to $-0.512$ (2B L22, 85\%) to
null at L24 (92\%), and from $-0.546$ (9B L20, 48\%) to $-0.327$ (9B L36, 86\%).
The geometry--steerability signal is strongest at shallow-to-mid proportional depth
and degrades as the intervention layer approaches the output, marking a practical
boundary for the framework.
 
\begin{figure}[t]
  \centering
  \includegraphics[width=\columnwidth]{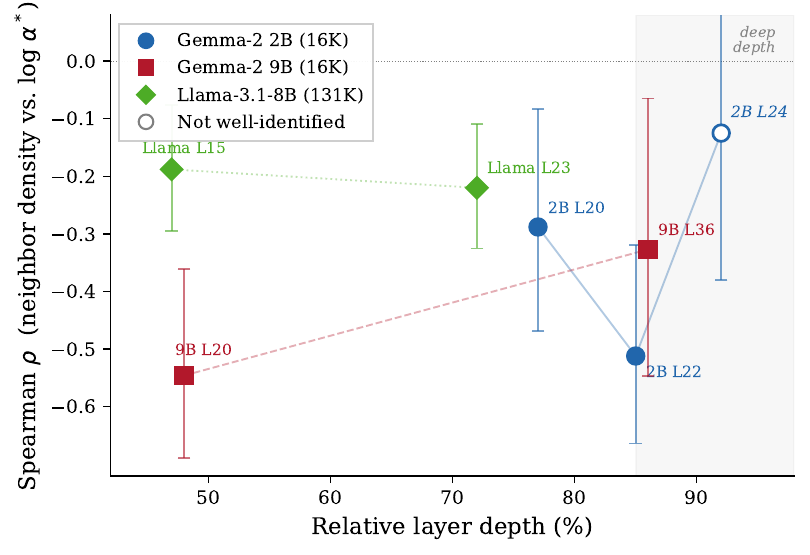}
  \caption{Spearman $\rho$ (neighbor density vs.\ $\log \alpha^*$) as a function
  of relative layer depth across all conditions. The geometry--steerability signal
  is strongest at shallow-to-mid depth and weakens consistently as the intervention
  layer approaches the output. Open marker indicates a not-well-identified condition
  (Section~\ref{sec:wellidentified}). Error bars show approximate 95\% CIs via
  Fisher $z$-transform. Shaded region marks the deep-depth boundary.}
  \label{fig:depth_boundary}
\end{figure}
 
\section{Conclusion}
\label{sec:conclusion}
 
Decoder-space geometry partially predicts SAE feature steerability before any
intervention is applied. Neighbor density and maximum cosine similarity consistently
rank-order features by $\log \alpha^*(f)$ across all well-identified conditions
($\rho$ up to $-0.546$, AUROC $0.822$ at 9B layer 20), the signal is detectable
cross-architecturally on Llama-3.1-8B-Instruct, and co-activation correlation
carries no independent predictive signal once activation sparsity is accounted for.
The framework has clear boundary conditions: the signal degrades at deep
proportional layer depth, and on Qwen3-8B with BatchTopK SAEs geometry predicts
binary steerability but not the continuous ordering among responsive features.
 
A practitioner with access to the SAE decoder weight matrix can screen features for
controllability before running a single steering experiment. Whether geometrically
dense features also produce greater off-target effects (RQ2 and RQ3) is the most
important open question for translating this framework into operational safety
guidance.
 
\subsection{Limitations}
 
Our experiments cover four models and one behavioral domain (refusal); whether the
continuous geometry--steerability relationship holds beyond GemmaScope's JumpReLU
training regime remains open, as Qwen3-8B with BatchTopK SAEs shows only binary
steerability prediction. The fixed coefficient grid attenuates results at deep
layers (46\% censoring at 9B L36) and should be calibrated per layer in future
work. The keyword classifier under-counts soft refusals, meaning reported effect
sizes are likely conservative (Appendix~\ref{app:scorer}). Co-activation
correlation is unreliable as operationalized due to zero-inflation. All experiments
use positive steering only; suppression ($\alpha < 0$) and the Llama generalization
(single layer, single run) are left to future work.
 
\subsection{Future Work}

Future work should test whether the same geometric metrics that predict steering
cost also predict off-target behavioral change --- evaluating features both at
their minimum effective coefficient $\alpha^*(f)$ and at a shared low coefficient
$\alpha_0 = 1$. A second extension would learn practical coefficient caps for
features in high-risk geometric neighborhoods. Formal definitions of the off-target
quantities are given in Appendix~\ref{app:offtarget}.

\subsection*{Broader Impact}
\label{sec:broader_impact}
 
The ability to predict which features are easy to steer could lower the barrier for
targeted misuse of activation-space interventions, particularly for refusal
suppression in safety-tuned models. We acknowledge this dual-use risk explicitly.
We note, however, that the same framework is directly useful defensively:
identifying geometrically isolated features that require large coefficients to steer, 
as well as crowded features that may produce off-target effects through neighbor 
co-recruitment, provides a tool for auditing steering pipelines
before deployment in safety-critical settings. Concretely, a safety team could use
pre-steering geometry to flag high-risk features --- those in dense neighborhoods
that respond at small coefficients and may produce collateral effects through 
neighbor co-recruitment ---
before any intervention is attempted, rather than discovering these properties
through expensive post-hoc evaluation. A framework that can predict intervention
difficulty and collateral risk in advance is more likely to surface unsafe steering
configurations than one that leaves these properties opaque. Responsible disclosure
practices apply; we will make code available contingent on safety review before
publication.

\appendix
\section{Feature Selection Details}
\label{app:selection}
 
For each SAE feature $f$, the composite contrast score is defined as
\begin{equation}
    C(f) = \text{act\_freq}_{\text{task}}(f) -
    \text{act\_freq}_{\text{neutral}}(f),
\end{equation}
where activation frequency is computed with max pooling over token positions within
each prompt. Features are ranked by $C(f)$ and the top 300 are retained.
 
For each feature $f$ in the contrast-selected set, we then compute an output
relevance score:
\begin{equation}
    \delta(f) = \bigl\| p_{\text{steered}}(\cdot \mid x) -
    p_{\text{base}}(\cdot \mid x) \bigr\|_1,
\end{equation}
averaged over a held-out prompt set, where $p_{\text{steered}}$ denotes the output
distribution with feature $f$ clamped at a reference coefficient. Features with
$\delta(f) < 0.01$ are discarded.
 
\section{Steering Grid and Censoring}
\label{app:grid}
 
The coefficient grid $\alpha \in \{0, 0.25, 0.5, 1, 2, 3, 5, 10, 20\}$ is chosen
to span two orders of magnitude with denser coverage at low coefficients, where
behavioral transitions are expected. The extended upper range (10, 20) is included
to capture hard-to-steer features that do not respond at lower coefficients; a
reference coefficient of $\alpha_{\text{ref}} = -20$ is used for baseline
estimation.
 
Right-censored features ($\alpha^*(f) > \alpha_{\max}$) are included in Spearman
correlations using standard rank-based handling of ties at the boundary and excluded
from $R^2$ regression to avoid artificial ceiling effects. We report censoring rates
alongside all summary statistics. Censoring rates varied substantially across
conditions, from 2.7\% (9B layer 20) to 46\% (9B layer 36), reflecting the
depth-dependent growth in steering cost described in Section~\ref{sec:results}.
 
\section{Refusal Scorer}
\label{app:scorer}
 
The on-target refusal classifier uses a keyword-matching approach: a completion is
labeled as a refusal if it contains any of a curated set of refusal phrases
(e.g., ``I cannot,'' ``I'm unable,'' ``As an AI''). The phrase list is provided in
the codebase (\texttt{src/refusal\_scorer.py}).
 
\paragraph{Classifier reliability.}
We evaluated the keyword classifier against an independent judge on 160 completions
stratified across $\alpha$ levels ($\alpha \in \{0, 0.5, 1, 2, 3, 5, 10, 20\}$),
with equal representation of refusal and non-refusal completions at each level.
 
Results: precision 97.9\%, recall 57.5\%, F1 72.4\%, overall agreement 78.1\%
($n = 160$; TP$=$46, TN$=$79, FP$=$1, FN$=$34). The high precision and moderate
recall reflect a systematic pattern: the classifier reliably identifies explicit
refusals but misses soft refusals --- completions that decline without using
canonical refusal phrases (e.g., ``this perpetuates a harmful stereotype'' without
an explicit ``I cannot''). Agreement degrades at high steering coefficients
($\alpha = 10$: 60\%; $\alpha = 20$: 50\%), where coherence collapse produces
degenerate repetition outputs that contain no refusal phrases and are scored 0 by
the classifier regardless of interpretability.
 
Because the classifier systematically under-counts refusals, our behavioral
threshold $T = 0.10$ is conservative: features that genuinely reach threshold may
be scored as not reaching it, meaning our reported $\alpha^*(f)$ values are likely
overestimates and our Spearman correlations are likely attenuated. The true
geometry--steerability relationship is therefore probably stronger than reported.

\section{Threshold Sensitivity}
\label{app:threshold}
 
We re-extract $\alpha^*(f)$ from existing $B(\alpha, f)$ curves for thresholds
$T \in \{0.05, 0.10, 0.15, 0.20\}$ without any new model runs. As $T$ increases,
more features are right-censored: for the 65K condition, censoring rises from 3\%
at $T=0.05$ to 30\% at $T=0.20$, confirming that $T=0.10$ sits in a
well-identified range. Only $\alpha^*(f)$ and the subsequent Spearman computation
change across columns; decoder geometry is fixed. Spearman correlations are stable
across all four thresholds, confirming $T = 0.10$ is not a cherry-picked boundary.
 
\section{Pre-Steering Geometry Statistics}
\label{app:geom_stats}
 
Table~\ref{tab:geom_stats} reports mean and standard deviation of neighbor density
and maximum cosine similarity across all conditions. The 65K SAE shows higher
absolute geometry values by construction of the wider dictionary. Llama-3.1-8B
geometry values are substantially higher than GemmaScope conditions, consistent
with the larger feature dictionary compressing decoder directions into a more
crowded space. Qwen3-8B geometry variance is notably compressed, consistent with
the BatchTopK training regime producing a more uniform decoder geometry.
 
\begin{table}[h]
\caption{Pre-steering geometry statistics per condition.}
\label{tab:geom_stats}
\vskip 0.1in
\begin{center}
\begin{small}
\begin{tabular}{lcccc}
\toprule
Condition & ND mean & ND std & MC mean & MC std \\
\midrule
2B L20 16K  & 0.194 & 0.053 & 0.392 & 0.143 \\
2B L22 16K  & 0.178 & 0.049 & 0.368 & 0.141 \\
2B L20 65K  & 0.323 & 0.078 & 0.547 & 0.139 \\
9B L20 16K  & 0.212 & 0.045 & 0.400 & 0.110 \\
9B L36 16K  & 0.180 & 0.045 & 0.373 & 0.127 \\
\bottomrule
\end{tabular}
\end{small}
\end{center}
{\footnotesize ND = neighbor density; MC = max cosine similarity.}
\vskip -0.1in
\end{table}
 
\section{Qwen3-8B Results}
\label{app:qwen}
 
We evaluate the geometry--steerability relationship on Qwen3-8B using a BatchTopK
SAE at layer 18 (${\sim}50\%$ relative depth). Across 113 contrast-selected
features with a baseline refusal rate of 0.46, we find a statistically significant
positive correlation between geometry metrics and $\log \alpha^*(f)$ for both max
cosine similarity ($\rho = 0.279$, $p = 0.003$) and neighbor density ($\rho =
0.275$, $p = 0.003$) on the full sample. However, restricting to the 98 uncensored
features, both correlations collapse to near zero ($\rho \approx 0.01$), indicating
that geometry predicts whether a feature is steerable at all rather than the
gradient of steerability among responsive features. The positive full-sample
correlation reflects that censored features (those that fall on the higher-geometry end of the 
distribution in this condition, consistent with a sign-reversed boundary 
case) dominate the sample; among uncensored features $\rho \approx 0.01$,
which is not a reversal of direction but an absence of continuous signal. This is
consistent with BatchTopK training combined with aggressive attention-sink filtering
compressing decoder variance and weakening the density--isolation contrast that the
framework relies on. Co-activation correlation is null ($\rho = 0.065$,
$p = 0.49$), consistent with all other conditions.
 
We attribute the qualitatively different steerability response to the BatchTopK
training regime. BatchTopK enforces sparsity by selecting the top-$k$ activations
across the entire batch rather than applying per-token thresholds as in GemmaScope's
JumpReLU training. This batch-level competition means features do not need to occupy
geometrically distinct, well-separated directions to remain active, decoupling
decoder geometry from functional behavior. Additionally, the Qwen SAE training
included aggressive filtering of attention sink activations, which are precisely
the high-norm, geometrically isolated activations that would otherwise create strong
directional structure in the decoder. Together, these choices appear to produce a
more uniform decoder geometry with compressed variance, reducing the signal
available for continuous geometry-based prediction.
 
The Qwen result therefore characterizes a boundary condition of the framework: the
geometry--steerability relationship as operationalized here applies to SAEs where
decoder geometry reflects functional isolation, and is attenuated or qualitatively
changed when SAE training decouples geometry from behavior. This is a useful
constraint on the generalizability of the framework.

 \section{Off-Target Analysis Definitions}
\label{app:offtarget}

The planned off-target analysis evaluates behavioral change in two regimes: at
$\alpha = \alpha^*(f)$ (risk at minimum effective push) and at a fixed low
coefficient $\alpha_0 = 1$ (testing whether features can be risky even when gently
steered). For off-target benchmarks $\{U_j\}_{j=1}^m$, the relevant quantities are:
\begin{align}
    R_{\text{mag}}(\alpha, f) &= \frac{1}{m}\sum_{j=1}^{m}
        |\Delta U_j(\alpha, f)|, \\
    R_{\text{breadth}}(\alpha, f) &= \frac{1}{m}\sum_{j=1}^{m}
        \mathbf{1}\bigl\{|\Delta U_j(\alpha, f)| > \tau\bigr\}.
\end{align}
Candidate off-target benchmarks include GPQA \citep{rein2023gpqa} for reasoning
and TruthfulQA \citep{lin2022truthfulqa} for truthfulness.
\end{document}